\documentclass[nohyperref]{article}

\usepackage{microtype}
\usepackage[dvipsnames]{xcolor}
\usepackage{graphicx}
\usepackage{subfigure}
\usepackage{booktabs} 
\usepackage{multirow}
\usepackage{tikz}
\usepackage{caption}
\usepackage{natbib}

\usepackage{hyperref}

\usepackage[accepted]{icml2022}

\usepackage{amsmath}
\usepackage{amssymb}
\usepackage{mathtools}
\usepackage{amsthm}

\theoremstyle{plain}

\theoremstyle{definition}

\theoremstyle{remark}

\newcommand{\R}{\mathbb{R}}
\newcommand{\E}{\mathbb{E}}
\newcommand{\bb}{\mathbf{b}}
\newcommand{\xxt}{\widetilde{\mathbf{x}}}
\newcommand{\xx}{\mathbf{x}}

\DeclareMathOperator*{\argmin}{arg\,min}

\icmltitlerunning{Towards Neural Sparse Linear Solvers}

\begin{document}

\twocolumn[
\icmltitle{Towards Neural Sparse Linear Solvers}



\icmlsetsymbol{equal}{*}

\begin{icmlauthorlist}
\icmlauthor{Luca Grementieri}{zuru}
\icmlauthor{Paolo Galeone}{zuru}
\end{icmlauthorlist}

\icmlaffiliation{zuru}{ZURU Tech, Modena, Italy}

\icmlcorrespondingauthor{Luca Grementieri}{luca.g@zuru.tech}

\icmlkeywords{sparse linear solvers, graph neural network, numerical analysis, }

\vskip 0.3in
]



\printAffiliationsAndNotice{}  

\begin{abstract}
Large sparse symmetric linear systems appear in several branches of science and engineering thanks to the widespread use of the finite element method (FEM). The fastest sparse linear solvers available implement hybrid iterative methods. These methods are based on heuristic algorithms to permute rows and columns or find a preconditioner matrix. In addition, they are inherently sequential, making them unable to leverage the GPU processing power entirely.
We propose neural sparse linear solvers, a deep learning framework to learn approximate solvers for sparse symmetric linear systems. Our method relies on representing a sparse symmetric linear system as an undirected weighted graph. Such graph representation is inherently permutation-equivariant and scale-invariant, and it can become the input to a graph neural network trained to regress the solution.
We test neural sparse linear solvers on static linear analysis problems from structural engineering.
Our method is less accurate than classic algorithms, but it is hardware-independent, fast on GPUs, and applicable to generic sparse symmetric systems without any additional hypothesis. Although many limitations remain, this study shows a general approach to tackle problems involving sparse symmetric matrices using graph neural networks.
\end{abstract}

\section{Introduction}
\label{introduction}

The resolution of linear systems is a classical problem at the core of the numerical analysis.
Linear systems are so pervasive because they are a fundamental component of the numerical resolution of
differential equations. For this reason, they appear in all branches of science: from engineering to biology.
The finite element method (FEM) is a popular and accurate method to solve differential equations numerically \citep{logan2016first}.
In this context, the matrix associated with the linear system has the property to be very sparse and symmetric (and often positive-definite).
The real-time simulation of FEM models requires fast resolution algorithms that can be inaccurate up to a certain degree.
Classic methods to solve sparse linear systems have been conceived before the introduction of GPGPU computing when the processing power available was very limited, so they are not suited to use a GPU effectively.
The methods to solve sparse linear systems are classified into direct and iterative methods.

Direct methods,  also known as decomposition methods, rely on a factorization of the coefficient matrix in a pair of matrices that
have simplifying properties, like triangular or orthogonal matrices.
The most established direct methods are LU factorization (equivalent to Cholesky decomposition for definite-positive matrices) and
QR decomposition.
Decomposition methods are very accurate, but their parallelization on modern hardware (GPUs) is complicated, and it is impossible to trade off accuracy for computational time.
Moreover, the factors obtained in the decomposition process can be much less sparse than the starting matrix, causing an increase in memory consumption.
Every null element of the initial matrix that becomes a non-null value during the factorization is called a \textit{fill-in}. When the number of fill-ins is large, the execution time increases considerably because the mutation of the sparsity pattern of a sparse matrix is a slow operation.
A permutation of rows and columns can mitigate the appearance of fill-ins by concentrating the non-null elements of the sparse matrix around the diagonal. 
The research of the optimal permutation that minimizes the number of fill-ins is an NP-hard problem \citep{yannakakis1981computing}, so heuristic fill-reducing algorithms are employed in practice.

Iterative methods repeatedly refine an approximate solution starting from an initial guess, often the null vector. Iterative methods have the advantage that a less stringent stopping criterion permits the reduction of the execution time at the expense of a less accurate solution. The coefficient matrix is never changed or inverted, but it is used only to perform matrix-vector multiplications. For this reason, iterations are fast on GPU but, since these methods are inherently sequential, the GPU usage is low, and the degree of parallelization is limited.
When the condition number of the coefficient matrix is high, iterative methods can incur a slow convergence increasing the total computation time. Preconditioning methods are harnessed to reduce the condition number, but they can be slow on GPUs because they often rely on decomposition methods.
The most adopted iterative methods are Conjugate Gradient (CG) or its variant Preconditioned Conjugate Gradient (PCG).
Such algorithms are only applicable to positive-definite matrices. When the coefficient matrix is indefinite,
it is possible to employ BiCGSTAB \citep{van1992bi} or GMRES \citep{saad1986gmres}, but they are often slower than CG
on systems of similar size.

We are interested in fast approximate solvers that can be used in real-time non-critical applications to predict a coarse solution of a linear system.
We present a deep learning framework to learn approximate sparse linear solvers tailored for a specific use case.
Our approximate solvers hinge on the recent advances in graph neural networks (GNNs) and geometric deep learning \citep{bronstein2017geometric}. To leverage GNNs, we present a novel way to represent a sparse symmetric linear system as a weighted undirected graph. Using this input representation, we can cast the resolution of a sparse linear system as a node regression task. We tackle this regression task using a graph convolution network (GCN) coupled with some techniques to exploit the scale invariance property of the problem.

We test our neural sparse linear solvers (NSLSs) on a dataset of synthetic and realistic symmetric positive-definite linear systems coming from the field of structural engineering.
Our tests show that our approximate solver is less accurate than existing methods and, for this reason, slower if compared with other iterative methods (being the relative error equal). Nevertheless, an NSLS has several advantages: 
it does not suffer from convergence issues caused by ill-conditioned matrices;
its implementation is hardware-independent;
the training on positive-definite, indefinite, or general sparse symmetric systems does not require any change.

Lastly, we consider our proposed deep learning approach an initial significant example of how deep learning can help solve
numerical analysis problems on sparse matrices. Our technique is part of the broader line of research toward 
computational-aided mathematics \citep{lample2019deep, davies2021advancing}.
 
To summarize, our contributions are the following:
\begin{itemize}
\item we introduce a novel representation of a sparse symmetric linear system as a weighted undirected graph;
\item we frame the resolution of a sparse linear system as a node regression task;
\item we build a GNN-based deep learning model to solve approximately linear systems.
\end{itemize}

\section{Background}
\label{background}

\subsection{Problem Definition}
Let $Sym_n$  be the set of real symmetric $n \times n$ matrices.
A symmetric linear system is defined by a symmetric matrix of coefficients
$A \in Sym_n$ and the vector of constant terms $\bb \in \R^n$.
The solution of the linear system is the vector $\xx \in \R^n$ such that
\begin{equation}
\label{eq:system}
A\xx = \bb.
\end{equation}
A system is said \textit{sparse} if $A$ is a sparse matrix, i.e.
the number of non-zero entries scales linearly with the number of rows 
and columns.

If $\det(A) \neq 0$, the linear system \eqref{eq:system} has a unique solution,
so it exists a solution function $s$ that maps the linear system to its solution
\begin{equation}
s(A, \bb) = A^{-1}\bb = \xx.
\end{equation}
$s$ is linear with respect to $\bb$, but it is non-linear with respect to $A$
because matrix inversion is a non-linear operation.

The computation of $s$ is slow when $n$ is large, so we look for an algorithm to approximate the function $s$.

\subsection{Preconditioning}
One possibility is to approximate the inversion of $A$ separating the linear part of $s$, the multiplication with $\bb$ from the non-linear inversion of $A$.
The line of research that proposes new preconditioning methods follows this approach implicitly.

A \textit{preconditioner} $M$ of a matrix $A$ is a matrix such that $M^{-1}A$ has a smaller condition number than $A$ (and $M^{-1}$ is easy to compute). 
When a preconditioner $M$ is available, the solution $\xx$ is computed solving the equivalent system $M^{-1}A \xx = M^{-1}\bb$.
For a symmetric matrix $A$ (a special case of a normal matrix), the condition number is $$\kappa(A) = ||A^{-1}|| \cdot ||A|| = \frac{|\lambda_{\text{max}}(A)|}{|\lambda_{\text{min}}({A})|} \geq 1,$$ where $\lambda_{\text{max}}(A)$ and $\lambda_{\text{min}}(A)$ are maximal and minimal (by modulus) eigenvalues of $A$ respectively. It follows that the minimum condition number is attained on multiples of the identity matrix.

Preconditioning methods try to approximate the inverse of $A$ under some restrictive conditions to minimize the condition number.
For example, usually, the sparsity pattern of $M$ or $M^{-1}$ coincides with the sparsity pattern of $A$.
With such limitation, the inverse $A^{-1}$ can be approximated but not recovered because the inverse matrix of a sparse matrix is non-sparse in the general case.
The methods that learn the preconditioning method based on a dataset of matrices often impose the same sparsity constraint \citep{gotz2018machine}.

\section{Graph Representation of a Sparse System}
\label{sec:graph}
In this work, we propose a method to approximate the function $s$ directly using a neural network.
Before applying any model, we have to represent the sparse system in a form appropriate to input it into a neural network model.

\citet{rose1972graph} shows that a sparsity profile of a symmetric matrix can be described by an undirected graph between $n$ nodes.
Starting from this idea, we define the graph $\mathcal{G}$ associated with a symmetric sparse linear system.

An \textit{undirected graph} $\mathcal{G}$ is a pair $(V, E)$ composed of a finite set of nodes $V$ and a
set of undirected edges $E \subseteq \{\{u, v\} \subseteq V\}$. A \textit{weighted graph} additionally associates to every edge $e \in E$ a real value $w(e)$, called its \textit{weight}. The \textit{adjacency matrix} of a weighted undirected graph is a symmetric matrix $G \in Sym_n$ such that $g_{ij} = g_{ji} = w(\{i, j\})$ if $\{i, j\} \in E$, otherwise $g_{ij} = 0$. Thus, we can interpret any symmetric sparse matrix as the adjacency matrix of a weighted undirected graph $\mathcal{G}$.

\begin{figure}[htb]
\begin{equation*}
\begin{pmatrix}
1.0 & 0.5 & 0 & 0 & 0 \\
0.5 & 2.2 & 4.1 & 0 & 1.2 \\
0 & 4.1 & -1.5 & 2.0 & 0 \\
0 & 0 & 2.0 & 3.6 & -0.8 \\
0 & 1.2 & 0 & -0.8 & -0.1
\end{pmatrix}
\begin{pmatrix}
x_1 \\
x_2 \\
x_3 \\
x_4 \\
x_5 
\end{pmatrix} 
= 
\begin{pmatrix}
\textcolor{BrickRed}{2.7} \\
\textcolor{orange}{-1.1} \\
\textcolor{ForestGreen}{-2.6} \\
\textcolor{NavyBlue}{5.4} \\
\textcolor{violet}{4.8}
\end{pmatrix} 
\end{equation*}
$$
\big\downarrow
$$
\begin{center}
\begin{tikzpicture}[auto,node distance=2cm,thick,main node/.style={circle,draw,minimum size=1.2cm,font=\sffamily}]]
\tikzset{every loop/.style={}}
\node[main node, draw=BrickRed] (1) {\textcolor{BrickRed}{$2.7$}};
\node[main node, draw=orange] (2) [right of=1] {\textcolor{orange}{$-1.1$}};
\node[main node, draw=ForestGreen] (3) [above right of=2] {\textcolor{ForestGreen}{$-2.6$}};
\node[main node, draw=NavyBlue] (4) [below right of=3] {\textcolor{NavyBlue}{$5.4$}};
\node[main node, draw=violet] (5) [below right of=2] {\textcolor{violet}{$4.8$}};

\path[every node/.style={font=\sffamily\small}]
    (1) edge node {$0.5$} (2)
        edge [loop above] node {$1.0$} (1)
    (2) edge node {$4.1$} (3)
        edge node {$1.2$} (5)
        edge [loop below] node {$2.2$} (2)
    (3) edge node {$2.0$} (4)
        edge [loop right] node {$-1.5$} (3)
    (4) edge [loop right] node {$3.6$} (4)
    (5) edge node {$-0.8$} (4)
        edge [loop right] node {$-0.1$} (5);
\end{tikzpicture}
\caption{Conversion from the matrix representation to the graph representation of a sparse symmetric linear system.}
\label{fig:graph}
\end{center}
\end{figure}
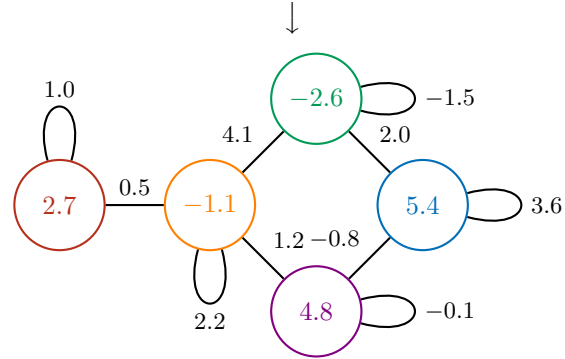

In our graph representation of a sparse linear system, the coefficients in $A$ become the weights of the undirected edges of the graph $\mathcal{G}$ and every node of $\mathcal{G}$ represent at the same time a variable of the system and one of its equations.
The order of equations in a linear system does not influence its solution, so there is no canonical ordering in general. Nevertheless, when the matrix is symmetric, the order of equations (i.e. the order of rows in the matrix) and the order of variables (i.e. the order of columns in the matrix) are linked by the symmetry property of the matrix. 
To complete the representation of the system as a graph, we use the elements of $\textbf{b}$ as node features.
In summary, as depicted in Figure \ref{fig:graph},
the $i^{\text{th}}$ node of the graph $\mathcal{G}$ is described by the feature $b_i$ and it is linked to the $j^{\text{th}}$ node if and only if $a_{ij} \neq 0$. If $a_{ij} \neq 0$, then the undirected edge $e_{ij}$ is associated to the weight $a_{ij}$. 

Apart from being more suitable to apply deep learning methods to the resolution of the problem, the graph representation is also inherently permutation-equivariant. It implies that an identical graph representation encodes symmetric systems obtained through permutations. This property is highly desirable because, if $P$ is a permutation matrix, then the system $A\mathbf{x}=\mathbf{b}$ is equivalent to the system
\begin{equation*}
(PAP^{\top})\mathbf{y}= P\mathbf{b}, \text{ where } \mathbf{y}=P \mathbf{x} \text{ is a permutation of } \textbf{x}.
\end{equation*}
On the contrary, the matrix representation of a linear system is not unique, and the order of rows and columns influences the resolution time of the linear system. For this reason, permutation heuristics try to find the most efficient form of a linear system. 

Linear systems are also scale-invariant: in Section \ref{ssec:scale}, we are
going to explain how to scale edge weights and node features to represent
scaled systems using the same graph.

\section{Neural Sparse Solver}
\begin{figure*}[thb]
\centering
\includegraphics[width=0.95\textwidth]{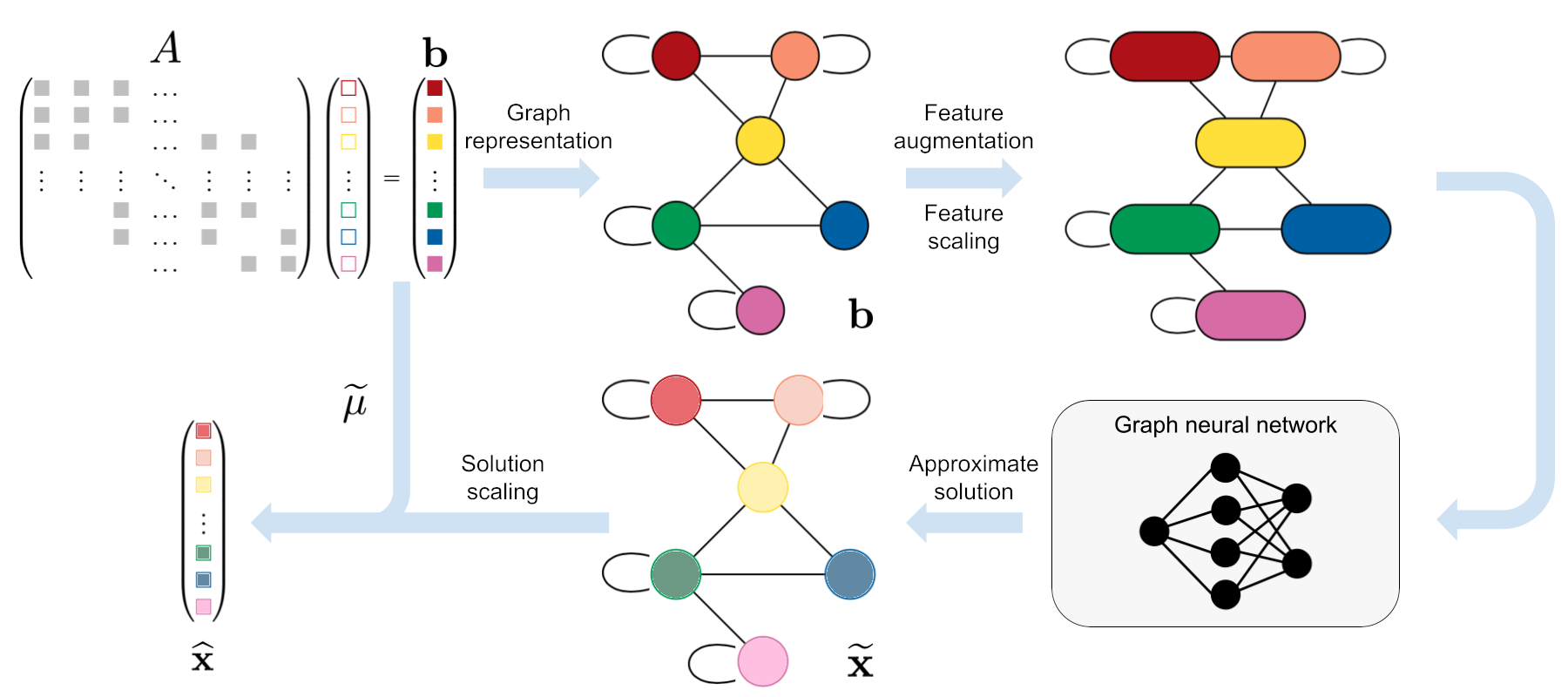}
\caption{Structure of a Neural Sparse Linear Solver. The system is converted to a graph: the coefficient matrix defines the graph topology, and the constant term $\bb$ gives the node features. Node features are augmented with permutation-equivariant operations and scaled. A graph neural network takes this graph as input and predicts the direction of the approximate solution. Finally, the output of the graph neural network is scaled to minimize the residual vector.}
\label{fig:nsls}
\end{figure*}

Equipped with the graph representation of a sparse linear system, we can now define our Neural Sparse Linear Solver (NSLS) model. An NSLS is a regression model that takes as input the graph associated with a sparse linear system according to the representation introduced in Section \ref{sec:graph}, and it outputs the approximate solution of the linear system.
Three modules compose an NSLS: a feature augmentation module, a scaling module, and a graph neural network module. Such modules interact according to the workflow displayed in Figure \ref{fig:nsls}.

\subsection{Permutation-Equivariant Feature Augmentation}

The naive application of a graph neural network on $\mathcal{G}_{A, \mathbf{b}}$ is prone to fail because every node stores a single scalar feature $b_i$. This description is insufficient to describe a node, so we need to enrich the node features and replace the scalar $b_i$ with the feature vector $\mathbf{f}_i$.

Every permutation-equivariant operation involving $A$ and $\textbf{b}$ that returns a vector $\mathbf{u} \in \mathbb{R}^n$, can be used to augment
the features of the $i^{\text{th}}$ node appending the value $u_i$ to it.
In practice, all features are normalized component-wise with the $L_{\infty}$ norm to control the variance of the input passed to the graph neural network. The $L_{\infty}$ norm is specifically chosen because it is independent of the size $n$.

The most simple augmentation is the \textit{diagonal augmentation}: the value $a_{ii}$ on the diagonal of the matrix $A$ is added to the feature vector $\mathbf{f}_i$. 
To design other augmentation operations, we draw inspiration from existing iterative methods.

\textbf{Jacobi Augmentation.} The Jacobi method is an iterative algorithm to solve the linear system $A \mathbf{x} = \mathbf{b}$ decomposing the matrix as $A = D + M$, where $D$ is the diagonal matrix, and $M$ is the sum of the lower and upper triangular parts. 
Jacobi iteration can be written as
\begin{equation*}
\begin{cases}
\mathbf{x}^{(0)}_{\text{Jacobi}} = \mathbf{0} \\
\mathbf{x}^{(k+1)}_{\text{Jacobi}} = D^{-1}(\mathbf{b}-M\mathbf{x}^{(k)}_{\text{Jacobi}}).
\end{cases}
\end{equation*}
We append to the feature matrix $F$ the vectors $\mathbf{x}^{(1)}_{\text{Jacobi}}, \ldots, \mathbf{x}^{(m)}_{\text{Jacobi}}$ after normalization. The normalization here is necessary because the Jacobi iteration can diverge rapidly if the matrix $A$ is not positive definite or strictly diagonally-dominant .

We could apply the same idea to similar iterative methods based on other decompositions of the matrix $A$, like Gauss-Seidel or SOR methods. Unfortunately, the iterations prescribed by these methods are not permutation-equivariant because the lower and upper parts of $A$ are different from the corresponding parts of the permuted matrix $PAP^{\top}$.

\textbf{Conjugate Gradient Augmentation.} A single iteration of the conjugate gradient method is a permutation-equivariant operation. Therefore, every conjugate gradient step provides values that can be appended element-wise to the corresponding node features. We expect this augmentation to be particularly effective when $A$ is positive-definite, but it is still possible to apply it to general matrices.

\textbf{Arnoldi Augmentation.} Arnoldi iteration is described by
\begin{equation*}
\mathbf{x}^{(k)}_{\text{Arnoldi}} = A^{k} \mathbf{b}.
\end{equation*}
The GMRES method \citep{saad1986gmres} is an iterative method for the numerical solution of a generic system of linear equations that relies on the Arnoldi iteration to approximate the solution. Thus, taking inspiration from GMRES, we enrich the node features appending the normalized version of the vectors $\mathbf{x}^{(1)}_{\text{Arnoldi}}, \ldots, \mathbf{x}^{(m)}_{\text{Arnoldi}}$. Here, the permutation equivariance property of the augmentation comes from the orthogonality of permutation matrices.
The normalization is crucial for Arnoldi augmentation because the repeated multiplication by $A$ can easily cause overflows or underflows in the values of $\mathbf{x}^{(k)}_{\text{Arnoldi}}$.

The proposed augmentations can be combined to maximize the accuracy of the model. The number of elements added to the node features should balance accuracy gains with the increase in inference time. 
We compare the end-to-end performances obtained using every augmentation strategy and their combinations in Section \ref{ssec:ablation}.

\subsection{Scale-Invariant Processing}
\label{ssec:scale}
The graph representation makes the model intrinsically permutation-equivariant: we now explain how to make the model scale-invariant. The scale invariance property allows us to describe entire classes of linear systems with a single representation, and it is even more critical for boosting model accuracy.
Indeed, inputs of uncontrolled scale can hinder the training or make the model unstable and unreliable.
Finally, scale invariance naturally helps generalization making it possible to manage solution vectors $\xx$ of different scales.

Let us notice that the system $A \xx = \bb$ is equivalent to
\begin{equation*}
\mu \frac{A}{||A||} \frac{\xx}{||\xx||} = \frac{\bb}{||\bb||}, \text{ where } 
\mu = \frac{||\xx|| ||A||}{||\bb||},
\end{equation*}
for any norm $|| \cdot ||$ (or in general for every scaling factor).
Let us denote the scaled quantities above as $\overline{A}, \overline{\xx}, \overline{\bb}$.
If the normalized solution $\overline{\xx}$ is known, we can recover 
\begin{equation*}
\mu = \frac{||\overline{\bb}||_2}{||\overline{A}\overline{\xx}||_2}.
\end{equation*}
If we only know an approximate scaled solution $\xxt$, then we can approximate $\mu$ as
\begin{equation*}
\widetilde{\mu} = \argmin_{\mu} ||\mu \overline{A} \xxt - \overline{\bb}||_2^{2}.
\end{equation*}
The minimization problem has the closed-form solution
\begin{equation}
\label{eq:approx_scale}
\widetilde{\mu} = \frac{\langle \overline{A} \xxt, \overline{\bb}\rangle}{||\overline{A} \xxt||_2^2}.
\end{equation}

In conclusion, we can represent all scaled versions of a system canonically using the normalized coefficients $\overline{A}$ and constant terms $\overline{\bb}$.
The model receives the system graph with scaled weights and features and, it predicts a scale-independent approximation of the solution $\xxt$.
Hence, we recover the final approximate solution $\widehat{\xx}$ with the formula
\begin{equation}
\widehat{\xx} = \frac{\widetilde{\mu} ||\bb||}{||A||} \xxt.
\end{equation}


\subsection{Graph Neural Network Backbone}
\begin{figure}[htb]
\centering
\includegraphics[width=0.4\linewidth]{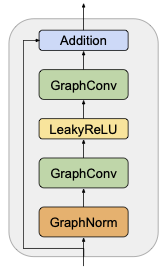}
\caption{Structure of a residual block in an NSLS.}
\label{fig:block}
\end{figure}

The main component of an NSLS is a graph convolutional neural network (GCN) that maps the scaled augmented node features to the approximate direction of the solution.
The GCN architecture follows the encode-process-decode model proposed by \citet{battaglia2018relational}: it is composed of a node encoder, followed by $m$ residual blocks, and terminates with a solution decoder.
The encoder is a linear layer, while the decoder is 2-layer MLP with LeakyReLU activation. The decoder outputs a single value per node, and all these values compose
a permutation-equivariant description of the solution vector.
For the residual block, we have tested many layers and their combinations to finally come up with the structure depicted in Figure \ref{fig:block}.

\textbf{Residual Connections.} \citet{li2020deepergcn} empirically find that the pre-activation variant of residual connections for GCNs, which
follows the ordering Normalization → Activation → GraphConv → Addition, performs better than the standard ordering introduced by \citet{li2019deepgcns}.
Here, we use a similar design with an additional GraphConv layer between the normalization and the activation layers. The supplementary convolutional layer increases the receptive field of the neural network earlier.

\textbf{Normalization Layer.} We have compared many normalization layers:
instance normalization \citep{ulyanov2016instance}, layer normalization \citep{ba2016layer}, graph normalization \citep{cai2021graphnorm}, graph size normalization \citep{dwivedi2020benchmarking}, pair normalization \citep{zhao2019pairnorm}. From our experiments, we see that graph normalization entails consistently the best performances. It is described by the formula
\begin{equation}
\label{eq:graphnorm}
\xx'_i = \frac{\xx - \alpha \cdot \E[\xx]}
{\sqrt{\textrm{Var}[\xx - \alpha \cdot \E[\xx]]
+ \epsilon}} \cdot \gamma + \beta, 
\end{equation}
where $\alpha, \beta, \gamma$ are scalar parameter learned during training.

\textbf{Convolution Layer.} The most common graph convolution layers, like the ones in GraphSAGE \citep{hamilton2017inductive}, GIN \citep{xu2018powerful} or GAT \citep{velivckovic2018graph}, are not suited for weighted graphs. Moreover, the convolutions in classic graph neural networks like GCN \citep{kipf2017semi} or GCNII
\citep{chen2020simple} can lead to zero division errors with some matrices.
Among the graph convolution operators which leverage edge weights, we adopt the GraphConv layer
proposed by \citet{morris2019weisfeiler}.
The update formula for the GraphConv layer is
\begin{equation}
\label{eq:graphconv}
\xx'_i = \Theta_1 \mathbf{x}_i + \Theta_2 \sum_{j \in \mathcal{N}(i)}a_{ji} \xx_j,
\end{equation}
where $\xx_i \in \R^d$ is the $i$-th node feature vector, $\Theta_1, \Theta_2 \in \R^{d' \times d}$ are the parameters of the layer, and $a_{ji}$ denotes the edge weight from source node $j$ to target node $i$.

Equation \eqref{eq:graphconv} shows that the message aggregation function is the sum. In the graph neural network literature many aggregation functions are advocated: for example, mean \citep{kipf2017semi}, max \citep{hamilton2017inductive}, softmax and power mean \citep{li2020deepergcn}.
Here we choose the sum aggregation function because, thanks to this choice, the message aggregation reduces to a matrix multiplication between the scaled coefficient matrix $\overline{A}$ and the processed node features $F \in \R^{n \times d}$. This choice is straightforward since the workhorse of many iterative methods is the matrix multiplication between the coefficient matrix and vectors depending on $\bb$.

\subsection{Loss Function}
The output of the GCN is a scale-independent vector, so we are just interested in its direction.
For this reason we train the model with the cosine distance loss function:
\begin{equation}
\mathcal{L}_{\text{cos}}(\xxt, \xx) =
1 - \frac{\langle \xxt, \xx \rangle}{||\xxt||_2 ||\xx||_2}.
\end{equation}

Surprisingly, this loss is not sufficient to recover a good approximation. We have observed that the vectors $\overline{A}\xxt$ and $\overline{b}$ can be near-orthogonal even if the angle between $\xxt$ and $\xx$ is tiny. This phenomenon happens because the transformation induced by an ill-conditioned matrix skews angles. To counteract this phenomenon, we use a complementary cosine distance loss on the vectors transformed by $\overline{A}$:
\begin{equation}
\mathcal{L}_{\text{res}}(\xxt, \xx) = 
\mathcal{L}_{\text{cos}}(\overline{A}\xxt, \overline{\bb}) =
1 - \frac{\langle \overline{A}\xxt, \overline{\bb} \rangle}{||\overline{A}\xxt||_2 ||\overline{\bb}||_2}.
\end{equation}
The minimization of $\mathcal{L}_{\text{res}}$ ensures that the projection of
$\overline{A}\xxt$ on $\overline{\bb}$ does not collapse into the null vector, making the estimation of $\widetilde{\mu}$ unreliable.

In conclusion, our full loss function is
\begin{equation}
\mathcal{L}(\xxt, \xx) = \mathcal{L}_{\text{cos}}(\xxt, \xx) + \mathcal{L}_{\text{res}}(\xxt, \xx).
\end{equation}

\section{Experiments}

\subsection{Results}

\textbf{Dataset.}
The most used collections of sparse matrices are the Matrix Market \citep{matrixmarket} and the SuiteSparse matrix collection \citep{suitesparsematrix}. They contain sparse matrices from many engineering problems, but these datasets are too small to train a deep learning algorithm on them. Furthermore, no constant term is reported in conjunction with these matrices. While simulating the solution is a viable option, it entails unrealistic linear systems. 

The only public dataset of symmetric sparse linear systems we have found is
the StAnD \citep{grementieri2022stand}. This dataset contains several solved linear systems originating in the stability analysis of a structure subjected to realistic loads. StAnD problems are divided by size: we perform our experiments on StAnD Small, the subset of linear systems with small coefficient matrices.
Besides, we also conduct our experiments on a synthetic dataset generated using the same matrices from StAnD Small but with solution elements uniformly sampled from $[-1, 1]$.
Our tests reveal very different behaviors induced by the two datasets, even if they share the same coefficient matrices. The elements in StAnD Small solution vectors range many orders of magnitudes (from $10^{-16}$ to $10^{-1}$), and we acknowledge this as the cause of the discrepancy.

The matrices in StAnD Small have an average dimension of 2115 rows and columns; this dimension corresponds to the number of nodes in the input system graph. Whereas the size of the system affects the execution time, the graph diameter conditions the precision of an NSLS: indeed, every node should collect information from every other node in its connected component. An NSLS can gather all values in $\bb$, necessary to produce the solution when the inverse of $A$ is dense, only if this condition holds. Otherwise, if the diameter of the system graph is larger than the receptive field of the message-passing neural network at the core of an NSLS, we know that the network has no theoretical possibility to produce a perfect solution. 
The receptive field of graph neural network depends on the connectivity of the input graph, but in the worst-case scenario of a linear graph 
it is equivalent to the number of convolution layers in the GCN.
The matrices in StAnD Small have an average diameter of about 20, so our models should have at least 20 convolution layers to be expressive enough. 

\textbf{Implementation.} All models are implemented using
PyTorch Geometric \citep{fey2019fast}. Our models have 10 residual blocks to guarantee a sufficiently wide receptive field since every residual block contains two GraphConv layers. 
The inner feature dimension $d$ is 32 (small) or 128 (medium), and the training lasts for 50 epochs, or about 315.000 steps using a batch size of 16.
We adopt the Adam optimizer \citep{kingma2014adam} with no additional regularization.
Indeed, we find that regularization is detrimental for such a regression task. 
We utilize a learning rate $\lambda=10^{-3}$ at the beginning of the training, dividing it by 10 after 40 and 45 epochs.

\textbf{Evaluation.} We evaluate our models computing on the test set the average of the mean absolute element-wise error $\epsilon$ and of the relative approximation error $\delta$ with respect to the ground-truth solution $\xx \in \R^n$ defined as
\begin{equation}
\epsilon(\widehat{\xx}, \xx) =  \frac{||\xx-\widehat{\xx}||_1}{n}, \qquad
\delta(\widehat{\xx}, \xx) = \frac{||\xx-\widehat{\xx}||_2}{||\xx||_2}.
\end{equation}
The relative error $\delta$ is scale-invariant, allowing us to compare models trained on different datasets. On the other hand, the mean absolute element-wise error $\epsilon$ tells us how far the approximation actually is from the solution. In practical applications, like structural engineering (the domain of StAnD), we are more interested in the absolute error because the values are associated with a unit of measure, and tiny errors are negligible. 

Table \ref{table:results} compares our leading models trained on synthetic and StAnD Small data. The feature augmentations applied to train such models are diagonal and conjugate gradient augmentations, with 14 steps of the conjugate gradient method to obtain feature vectors of size 16. This combination of features attains the best compromise between accuracy and speed.
The results on synthetic data are very satisfactory, exhibiting compelling stability even for the ill-conditioned matrices typical of structural analysis.
The relative error on actual StAnD data is much higher because the very low norm of the ground-truth solution distorts it. The absolute error $\epsilon$ tells us another story, showing that the approximation is adequately good and the NSLS is sufficiently accurate to replace a classic solver.

\begin{table}[ht]
\caption{Evaluation on synthetic and StAnD Small test sets.}
\label{table:results}
\vskip 0.15in
\begin{center}
\begin{small}
\begin{sc}
\begin{tabular}{llcc}
\toprule
Dataset & Inner Size & $\epsilon$ & $\delta$ \\
\midrule
\multirow{2}{*}{Synthetic} & $d = 32$ & $9.77 \cdot 10^{-3}$ & 3.40\% \\
& $d = 128$ & $8.03 \cdot 10^{-3}$ & 2.80\% \\
\midrule
\multirow{2}{*}{StAnD Small}& $d = 32$ & $8.27 \cdot 10^{-6}$ & 18.28\% \\
& $d = 128$ & $5.60 \cdot 10^{-6}$ & 10.55\%\\
\bottomrule
\end{tabular}
\end{sc}
\end{small}
\end{center}
\vskip -0.1in
\end{table}

\subsection{Ablation Study}
\label{ssec:ablation}
In Table \ref{table:ablation}, we present the results of the ablation study performed to verify the effectiveness of each module of an NSLS. 
To reduce the running time of experiments, we compare only models
with inner feature dimension $d = 32$ on StAnD Small.
Even if not specified, we always include the diagonal augmentation.

\begin{table}[ht]
\caption{Ablation study on StAnD Small test set.}
\label{table:ablation}
\vskip 0.15in
\begin{center}
\begin{small}
\begin{sc}
\begin{tabular}{llcc}
\toprule
Loss & Augmentation & $\epsilon$ & $\delta$ \\
\midrule
$\mathcal{L}_{\textup{cos}}$ & - & $18.3 \cdot 10^{-6}$ & 33.93\%  \\
$\mathcal{L}_{\textup{cos}} + \mathcal{L}_{\textup{res}}$ & - & $9.01 \cdot 10^{-6}$ & 21.66\%  \\
$\mathcal{L}_{\textup{cos}} + \mathcal{L}_{\textup{res}}$ & Arnoldi & $8.44 \cdot 10^{-6}$ & 19.86 \% \\
$\mathcal{L}_{\textup{cos}} + \mathcal{L}_{\textup{res}}$ & Jacobi & $8.89 \cdot 10^{-6}$ & 19.22\% \\
$\mathcal{L}_{\textup{cos}} + \mathcal{L}_{\textup{res}}$ & CG & $\mathbf{8.27 \cdot 10^{-6}}$ & 18.28\% \\
$\mathcal{L}_{\textup{cos}} + \mathcal{L}_{\textup{res}}$ & CG + Arnoldi & $8.53 \cdot 10^{-6}$ & 19.04\% \\
$\mathcal{L}_{\textup{cos}} + \mathcal{L}_{\textup{res}}$ & CG + Jacobi & $8.30 \cdot 10^{-6}$ & \textbf{18.12\%} \\
\bottomrule
\end{tabular}
\end{sc}
\end{small}
\end{center}
\vskip -0.1in
\end{table}

The metrics show that the $\mathcal{L}_{\text{res}}$ loss is a fundamental component of our architecture. Without this term, the model tends to overfit training data, and the gap between training loss and validation loss increases during the training.

The comparison between feature augmentation strategies shows that all proposed methods improve the final result. As expected, the conjugate gradient augmentation is the most beneficial on positive-definite matrices. On the other hand, Arnoldi iteration, which does not converge to the solution, is the least effective strategy, and it is also detrimental when combined with conjugate gradient.

\subsection{Comparison with Existing Methods}
We compare the running time of NSLS with the fastest open-source C++ libraries with GPU support for the resolution of linear systems: cuSOLVER \citep{cusolver} for direct methods and ViennaCL \citep{rupp2016viennacl} for iterative algorithms.
Since the matrices in StAnD have the additional property to be positive-definite, we compare NSLS with algorithms devised especially for positive-definite systems: Cholesky decomposition (cuSOLVER) and (Preconditioned) Conjugate Gradient (ViennaCL).
We remark that our method does not need the additional hypothesis of positive-definiteness.
On the contrary, we believe that neural solvers would be much more competitive on general symmetric systems that are harder to solve with classic algorithms. 
For reference, we also measure the execution time of a pure PyTorch implementation of the conjugate gradient to show that using the same framework and the same amount of user optimizations, NSLSs are faster.

We combine Cholesky decomposition with the fill-in reducing permutation strategies SymAMD \citep{amestoy1996approximate}, SymRCM \citep{george1971computer} or METIS \citep{karypis1998fast}.
Similarly, we consider Incomplete Cholesky \citep{lin1999incomplete} and Incomplete LU \citep{meijerink1977iterative} preconditioning to foster the convergence of the conjugate gradient method.

Direct methods cannot exchange accuracy for execution time, so they always
reach a very low error. For iterative methods, we record the time needed to reach a relative error equivalent to the one incurred by the NSLS models listed in Table \ref{table:results}. The execution time is measured on a GPU NVidia GeForce GTX 1080 Ti.
\begin{table}[ht]
\caption{Execution time needed by solvers to reach the average relative error $\delta$ of the best NSLS with $d= 32$ on synthetic and StAnD Small test data.}
\label{table:benchmark}
\vskip 0.08in
\begin{center}
\begin{small}
\begin{sc}
\resizebox{\columnwidth}{!}{
\begin{tabular}{llcc}
\toprule
\multirow{2}{*}{Algorithm} & \multirow{2}{*}{Variant} & \multirow{2}{*}{\shortstack{Synthetic\\Time (\textup{ms})}} 
& \multirow{2}{*}{\shortstack{StAnD Small\\Time (\textup{ms})}} \\
&&& \\
\midrule
\multirow{4}{*}{\shortstack{Cholesky\\(cuSPARSE)}} & - & 41.9 & 42.1\\
& SymAMD & 35.3 & 35.3 \\
& SymRCM & 36.1 & 36.7 \\
& METIS & 31.2 & 31.1 \\
\midrule
\multirow{3}{*}{\shortstack{Conjugate\\Gradient\\(ViennaCL)}} & - & 2.16 & \textbf{0.84} \\
& ICC & \textbf{1.97} & 1.20 \\
& ILU & 2.49 & 1.77 \\
\midrule
\multirow{3}{*}{\shortstack{Conjugate\\Gradient\\(PyTorch)}} & \multirow{3}{*}{-} & \multirow{3}{*}{38.1} & \multirow{3}{*}{14.4}\\
&&& \\
&&& \\
\midrule
\multirow{2}{*}{NSLS} & $d=32$ & 13.2 & 13.4\\
& $d=128$ & 18.7 & 18.5 \\
\bottomrule
\end{tabular}
}
\end{sc}
\end{small}
\end{center}
\vskip -0.1in
\end{table}

NSLSs are still slower than iterative methods, but they enjoy a better utilization of GPU resources and a higher degree of parallelization, as can be seen from the comparison with the pure PyTorch implementation of conjugate gradient.
Moreover, the implementation of NSLS in deep learning frameworks like PyTorch \citep{pytorch} or TensorFlow \citep{abadi2016tensorflow} makes these models hardware-agnostic and highly optimized without much effort.
We are sure that the inference time of neural solvers will be further reduced by the development of technology.
Many research papers already propose more optimized implementations of the sparse matrix multiplication on GPU \citep{huang2020ge}, and deep learning frameworks are actively adding support for advanced sparse formats and operations. Anyway, many burdens persist: for example, many neural compilers, like Apache TVM \citep{chen2018tvm}, do not support variable-size inputs, and thus the optimization of graph neural networks is challenging. Furthermore, the ONNX \citep{bai2019} support for sparse operation is very recent (see ONNX opset 16), and accelerators like ONNX Runtime \citep{onnxruntime} still have no support for such operations.

\subsection{Integration with Iterative Methods}
NSLSs provide a coarse approximation of the solution, so they are suited only for non-critical problems where the time constraints are more important than accuracy constraints, for example, in real-time simulations.
If more precision is needed, the output of an NSLS can furnish a high-quality initial guess to a classic iterative method.
Table \ref{table:iterations} shows that the NLSL initialization obtained with the medium model on StAnD Small speeds up the convergence of conjugate gradient, with a substantial reduction in the number of iterations.

\begin{table}[ht]
\caption{The average number of iterations needed by the conjugate gradient algorithm to reach a certain relative error on StAnd Small using different initialization vectors.}
\label{table:iterations}
\vskip 0.15in
\begin{center}
\begin{small}
\begin{sc}
\begin{tabular}{lccc}
\toprule
\multirow{2}{*}{\shortstack{Relative\\Error $\delta$}} &
\multirow{2}{*}{\shortstack{Zero Init\\\textup{(iterations)}}} &
\multirow{2}{*}{\shortstack{NSLS Init\\\textup{(iterations)}}} &
\multirow{2}{*}{\shortstack{Percentage\\Reduction}} \\
&&&\\
\midrule
1\% & 240 & 137 & -43\% \\
0.1\% & 362 & 270 & -25\% \\
0.01\% & 481 & 395 & -18\% \\
0.001\% & 583 & 512 & -12\% \\
\bottomrule
\end{tabular}
\end{sc}
\end{small}
\end{center}
\vskip -0.1in
\end{table}

\section{Related Works}

\textbf{Graph Representation.} \citet{parter1961use} first introduced explicitly the graph representation of a sparse matrix. \citet{rose1972graph} pursued a detailed graph-theoretic study of the graph representation for sparse symmetric positive-definite matrices. The graph representation of the matrix is still exploited to design permutation algorithms to reduce the number of fill-ins in Gaussian elimination. This line of work uses an unweighted graph, and nodes have no features since the graph represents just the matrix of coefficient, not the entire linear system.

\textbf{Linear System Solvers.} Most of the research on linear system solvers focuses on reduced classes of linear systems, like symmetric diagonally dominant (SDD) \citep{koutis2012fast, spielman2014nearly} or graph-structured \citep{kyng2016approximate} linear systems. Recently \citet{peng2021solving} proposed a method asymptotically faster than matrix multiplication applicable to the general case. This last algorithm is comprehensibly very complex, involving many steps and tricks. Unfortunately, no public implementation of the method is available, so it is impossible to test it and measure its actual running time.

\textbf{Preconditioners.} Recent preconditioning methods are specifically designed to be highly parallel and use efficiently GPUs \citep{anzt2018incomplete, dziekonski2018block, bernaschi2019dynamic, he2020efficient, gobel2021mixed}. Since no preconditioning algorithm works better than others on every matrix, some authors propose to use deep learning to generate the preconditioner matrix \citep{gotz2018machine, sappl2019deep, luna2021accelerating}. Preconditioning techniques complement our approach because they improve existing iterative methods, and we have seen how the integration of NSLSs into existing algorithms can reduce the number of iterations.

\section{Conclusion}

In this work, we present a deep learning model able to solve approximately symmetric sparse linear systems.
We test the method on sparse positive-definite linear systems emerging in the field of structural engineering.
On this task, the method is not very accurate and slower than classic methods, but the results obtained on a synthetic
dataset show the potential capabilities of the model. Moreover, our method suffers from the immaturity of the support
for sparse operations in deep learning frameworks. We are confident that the execution time of neural solvers will diminish
with further developments of the deep learning libraries.

Overall, our research can be considered a relevant application of a novel general approach to address numerical analysis problems on sparse matrices. For example, we could adapt it to predict an optimal preconditioner matrix for a linear system or an improved initialization for a specific iterative method.
Such examples of further developments show the generality of our approach and its flexibility.
Since the adaptation of algorithms to the sparse case is challenging and even more troublesome is to parallelize them to use GPUs, we believe that graph neural networks can be an effective alternative to current methods.

\bibliographystyle{icml2022}
\bibliography{nsls}

\end{document}